\documentclass{article}

\usepackage{booktabs} 

\usepackage{hyperref}

\usepackage[accepted]{sysml2019}

\usepackage{bibspacing}

\usepackage{authblk}
\usepackage{todonotes}
\usepackage{wrapfig}
\usepackage{caption} 
\usepackage{subcaption}
\usepackage[cal=boondox]{mathalfa}
\usepackage{marginnote}

\usepackage{mathrsfs}
\usepackage{amsmath}
\usepackage{amssymb}
\newcommand{\rom}[1]{%
  \textup{\uppercase\expandafter{\romannumeral#1}}%
}

\usepackage[utf8]{inputenc}
\usepackage{amsthm}
\usepackage{thmtools, thm-restate}

\newtheorem*{theorem*}{Theorem}

\usepackage{booktabs, adjustbox}
\usepackage{multirow}
\usepackage{threeparttable}
\usepackage{graphicx}
\usepackage{array}
\usepackage{tabularx}

\newcommand{\bx}{{\pmb x}}
\newcommand{\by}{{\pmb y}}
\newcommand{\bz}{{\pmb z}}

\newcommand{\be}{{\pmb e}}

\newcommand{\balpha}{{\pmb \alpha}}

\newcommand{\HH}{{\mathcal H}}

\let\oldphi\phi
\renewcommand\phi{\operatorname{\oldphi}}

\newcommand{\rme}{e}

\newcommand{\rmk}{k}
\newcommand{\rmH}{\mathcal{K}}

\newcommand{\rmP}{\mathcal{P}}

\newcommand{\TM}{\mathpzc{G}}

\newcommand{\KM}{K}
\newcommand{\mG}{m_{\TM}^{max}}
\newcommand{\epp}{q}
\newcommand\rx[1]{\bx_{r_#1}}
\newcommand\tx[1]{\bx_{t_#1}}
\newcommand\ty[1]{y_{t_#1}}

\newcommand{\tr}{\mathsf{tr}}

\newcommand\defeq{\triangleq}
\newcommand\norm[1]{\left\lVert#1\right\rVert}

\usepackage{mathtools}

\usepackage{mathrsfs}
\DeclareMathAlphabet{\mathpzc}{OT1}{pzc}{m}{it}
\newcommand\iprod[2]{\langle #1, #2 \rangle}

\newcommand\expect[1]{\mathbb{E}{\left\lbrack#1\right\rbrack}}

\theoremstyle{definition}
\newtheorem{remark}{Remark}[section]

\usepackage[toc,page]{appendix}

\usepackage{lipsum}

\usepackage{mathtools}
\usepackage{listings}
\usepackage{tikz}
\usepackage[nomessages]{fp}
\usepackage{adjustbox}
\usepackage{outlines}
\usepackage{times}
\usepackage{graphics}
\usepackage{array,float}
\usepackage{url}
\usepackage{amstext,amssymb,amsmath}
\usepackage{hyphenat}
\usepackage{amsthm}
\usepackage{verbatim}
\usepackage{bm}
\usepackage{paralist}
\usepackage{ulem}

\usepackage{etoolbox}

\newcommand{\surround}[2][r]%
  {\ifstrequal{#1}{round}%
    {\left( #2 \right)}%
    {\ifstrequal{#1}{square}%
      {\left[ #2 \right]}%
      {\ifstrequal{#1}{curly}%
        {\left\{ #2 \right\}}%
        {\ifstrequal{#1}{angle}%
          {\left\langle #2 \right\rangle}%
          {\ifstrequal{#1}{|}%
            {\left\lvert #2 \right\rvert}%
            {\ifstrequal{#1}{||}%
              {\left\lVert #2 \right\rVert}%
              {\ifstrequal{#1}{floor}%
                {\left\lfloor #2 \right\rfloor}%
                {\ifstrequal{#1}{ceil}%
                  {\left\lceil #2 \right\rceil}%
                  {\ifstrequal{#1}{.}%
                    {\left. #2 \right.}%
                    {\left( #2 \right)}%
                  }%
                }%
              }%
            }%
          }%
        }%
      }%
    }%
  }

\newcommand{\R}{\mathbb{R}}

\sysmltitlerunning{Kernel machines that adapt to GPUs for effective large batch training}

\begin{document}

\twocolumn[
\sysmltitle{Kernel machines that adapt to GPUs for \\effective large batch training}



\sysmlsetsymbol{equal}{*}

\begin{sysmlauthorlist}
\sysmlauthor{Siyuan Ma}{osu}
\sysmlauthor{Mikhail Belkin}{osu}
\end{sysmlauthorlist}

\sysmlaffiliation{osu}{Department of Computer Science and Engineering, Ohio State University, Columbus, Ohio, United States}

\sysmlcorrespondingauthor{Siyuan Ma~~~~~}{masi@cse.ohio-state.edu}
\sysmlcorrespondingauthor{~~~~~Mikhail Belkin}{mbelkin@cse.ohio-state.edu}

\sysmlkeywords{Machine Learning, SysML}

\vskip 0.3in

\begin{abstract}


Modern machine learning models are typically trained using Stochastic Gradient Descent (SGD) on massively parallel computing resources such as GPUs. Increasing mini-batch size is a simple and direct way to utilize the parallel computing capacity. 
For small batch an increase in batch size results in the proportional reduction in the training time, a phenomenon known as {\it linear scaling}. 
However, increasing batch size beyond a certain value leads to no further improvement in training time. In this paper we develop the first analytical framework that  extends linear scaling to match the parallel computing capacity of a resource.
The framework is designed for a class of classical kernel machines. It automatically modifies a standard kernel machine to output  a mathematically equivalent prediction function, yet allowing for  extended linear scaling, i.e., higher effective parallelization and faster  training time on given hardware.





The resulting algorithms are accurate, principled and very fast. For example, using a single Titan Xp GPU,  training on ImageNet with $1.3\times 10^6$ data points and $1000$ labels  takes under an hour, while smaller datasets, such as MNIST, take seconds. As the parameters are chosen analytically, based on the theoretical bounds, little tuning beyond  selecting the kernel and the kernel parameter is needed, further facilitating the practical use of these methods. 
\end{abstract}
]

\printAffiliationsAndNotice{~}

\section{Introduction}
\label{sec:intro}

Modern machine learning models are  trained using Stochastic Gradient Descent (SGD) on  parallel computing resources such as GPUs.
During training we aim to minimize the  (wall clock) training time $T_\text{train}$ given a computational resource, e.g., a bank of GPUs.
Although using larger batch size $m$ improves resource utilization, it does not necessarily lead to a reduction in training time. 
Indeed, we can decompose the training time $T_\text{train}(m)$ into two parts,
$$ 
T_\text{train}(m) = N_\text{epoch}(m) \times T_\text{epoch}(m) 
$$ 
where $N_\text{epoch}(m)$ is the number of training epochs required for convergence and $T_\text{epoch}(m)$ is the wall clock time to train for one epoch. 
It is easy to see that increasing $m$ always leads to higher resource utilization, thus decreasing $T_\text{epoch}(m)$.
However,  $N_\text{epoch}(m)$ may increase with $m$.
In fact, for a general class of convex problems it can be shown~\cite{ma2017interpolation} $N_\text{epoch}(m)$ is approximately constant for $m$ no more than a certain critical size $m^*$ and $N_\text{epoch}(m) \propto m$ for $m>m^*$.
On the other hand $T_\text{epoch}(m)$ is at best\footnote{Assuming perfect parallel computation.} decreases proportionally to $1/m$. 
Thus the training time is at least
$$
T_\text{train}(m)= \left\{\begin{matrix}
1/m, & \text{for~~} m \leq m^* \\ 
\text{const}, & \text{for~~} m > m^*
\end{matrix}\right.
$$


In other words, we obtain linear speedup (``linear scaling'') for batch sizes up to $m^*$, beyond which the training time cannot be improved by further increasing $m$. Furthermore, an important property of $m^*$ is its near independence from the number of training samples as it is primarily determined by the model and the data distribution.

Similar relationship between the batch size and the training time has been observed empirically in training deep neural networks~\cite{krizhevsky2014one}. A  heuristic called the ``linear scaling rule'' has been  widely used in deep learning practice~\cite{goyal2017accurate,you2017scaling,jia2018highly}.
Moreover, in parallel to the convex case analyzed in~\cite{ma2017interpolation}, recent work ~\cite{golmant2018computational,mccandlish2018empirical} empirically demonstrates that $m^*$ is independent of the data size for deep neural networks.


Many best practices of modern large-scale learning~\cite{goyal2017accurate,you2017scaling,jia2018highly} start with estimating $m^*$ by either heuristic rules or experiments. The optimal training time for the model is then capped by the estimated batch size $m^*$ which is  fixed  given the model architecture and weight, as well as the learning task.

In this work we propose a principled framework (EigenPro 2.0) that increases $m^*$ for a class of models corresponding to classical kernel machines. 
Our framework modifies a kernel machine to output a mathematically equivalent prediction function, yet allowing for extended linear scaling adaptive to (potentially) arbitrary parallel computational resource.
Furthermore, the optimization parameter selection is analytic, making it easy and efficient to use in practice and appropriate for ``interactive" exploratory machine learning and automatic model selection.
The resulting algorithms show significant speedup for training on GPUs over the state-of-the-art methods and excellent generalization performance.


\noindent{\bf Kernel machines.} 
Kernel machines are a powerful class of methods for classification and regression. Given the training data   $\{(\bx_i,y_i), i=1,\ldots,n\} \in \R^d\times \R$, and a positive definite kernel  $\rmk: \R^d\times \R^d \to \R$, kernel machines  construct  functions of the form  $f(\bx) = \sum_i \alpha_i \rmk(\bx,\bx_i)$. These methods are theoretically attractive,
show excellent performance on small datasets, and are known to be universal learners, i.e., capable of approximating any function from data. However, making kernel machines fast and scalable to large data has been a challenging problem.
Recent large scale efforts typically involved significant parallel computational resources, such as multiple (sometimes thousands) AWS vCPU's~\cite{tu2016large,avron2016faster} or super-computer nodes~\cite{huang2014kernel}. 
Very recently, FALKON~\cite{rudi2017falkon} and EigenPro~\cite{ma2017diving} showed strong classification results on large datasets with much lower computational requirements, a few hours on a single GPU.


\noindent{\bf The main problem and our contribution.} 
The main problem addressed in this paper is to minimize the training time for a kernel machine, given access to a parallel computational resource $\TM$. Our main contribution is that given a standard kernel, we are able to learn a new data and computational resource dependent kernel to minimize the resource time required for training without changing the mathematical  solution for the original kernel. 
Our model for a {\it computational resource} $\TM$ is based on a modern graphics processing unit (GPU), a device that allows for  very efficient, highly parallel\footnote{For example, there are 3840 CUDA cores in Nvidia GTX Titan Xp (Pascal).} matrix multiplication. 
\newline

The outline of our approach is shown in the diagram on the right. We now outline  the key ingredients.

\noindent{\bf The interpolation framework.} 
In recent years we have seen that inference methods, notably neural networks, that interpolate or nearly interpolate the training data generalize very well to test data~\cite{zhang2016understanding}. It has been observed in~\cite{belkin2018understand} that minimum norm kernel interpolants, i.e., functions of the forms $f(\bx) = \sum_i \alpha_i \rmk(\bx,\bx_i)$, such that $f(\bx_i) = y_i$, achieve optimal or 
near optimal generalization performance.  While the mathematical foundations of why interpolation produces good 
test results are not yet fully understood, the simplicity 
of the  framework can be used to   accelerate 
\begin{wrapfigure}{r}{0.24\textwidth}
  \includegraphics[width=.23\textwidth]{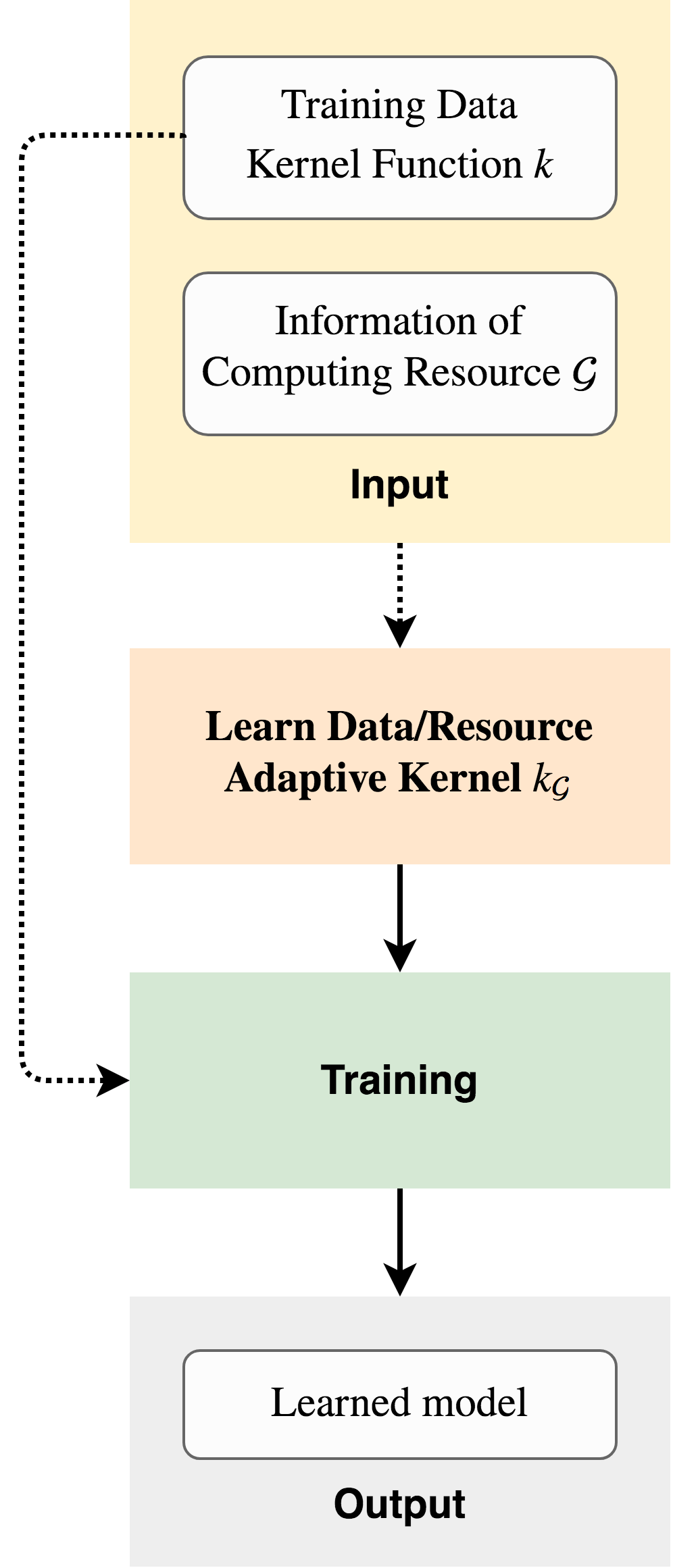}
  \label{fig:train}
\end{wrapfigure}
and scale the training of  classical kernel methods, while improving their test accuracy.
Indeed, constructing these interpolating functions is conceptually and mathematically simple, requiring approximately solving a single system of linear equations with a unique solution,  same for both regression and classification. Significant computational savings and, when necessary, regularization~\cite{yao2007early} are provided by early stopping, i.e., stopping iterations well before numerical convergence, once successive iterations fail to improve validation error. 

\noindent{\bf Adaptivity to data and computational resource: choosing optimal batch size and step size for SGD.} 
We will train kernel methods using Stochastic Gradient Descent (SGD), a method which is well-suited to modern GPU's and has shown impressive success in training neural networks. Importantly,  in the interpolation framework, dependence of convergence on the batch size and the 
step size can be derived analytically, allowing for full analysis  and automatic parameter selection.

We first note that in the parallel model each iteration of SGD (essentially a matrix multiplication) takes the same time for any mini-batch size up to $\mG$, defined as the mini-batch size where the  parallel capacity of the resource $\TM$ is fully utilized. It is shown in~\cite{ma2017interpolation} that in the interpolation framework convergence per iteration (using optimal step size) improves nearly linearly as a function of the mini-batch size $m$ up to a certain {\it critical size} $m^*(k)$ and rapidly saturates after that. The quantity $m^*(k)$ is related to the spectrum of the kernel.  For kernels used in practice it is typically quite small, less than $10$, due to their rapid eigenvalue decay. Yet, depending on the number of data points, features and labels, a modern GPU can  handle mini-batches of size $1000$ or larger. This disparity presents an opportunity for major improvements in the efficiency of kernel methods. 
In this paper we  show how to construct  data and resource adaptive kernel $k_\TM$, by modifying the spectrum of the kernel by using EigenPro algorithm~\cite{ma2017diving}.
The resulting iterative method with the new kernel has similar or better convergence per iteration than the original kernel $k$ for small mini-batch size. However its convergence improves linearly to much larger mini-batch sizes, matching  $\mG$, the maximum that can be utilized by the resource $\TM$. Importantly, SGD for either  kernel converge to the same interpolated solution. 

\begin{figure}[!ht]
  \vspace{-4mm}
  \includegraphics[width=.47\textwidth]{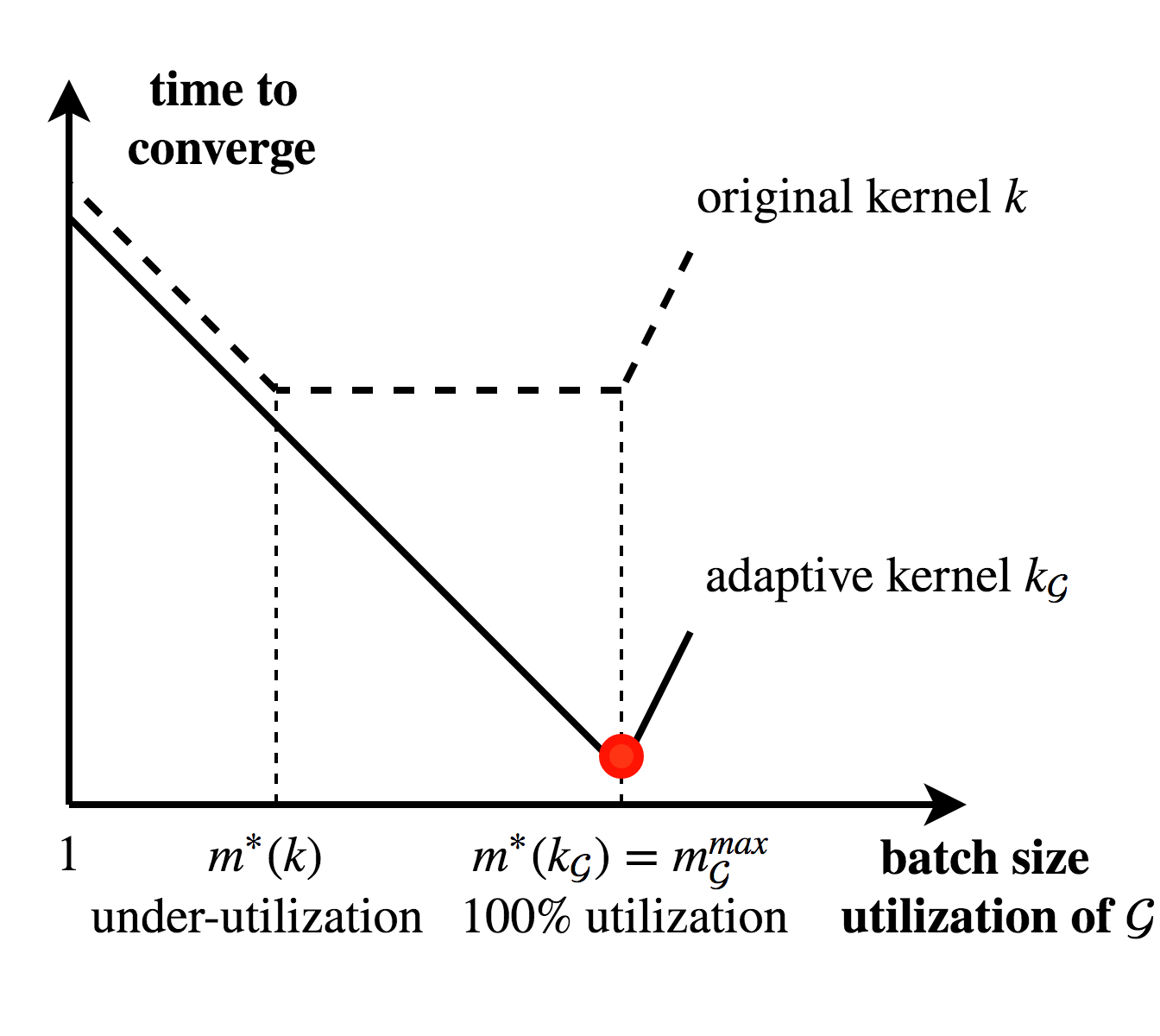}
    \vspace{-5mm}
\caption{Adaptive and original kernel}
  \label{fig:adaptive_vs_original}
\end{figure}
Thus, we aim to modify the kernel by constructing a kernel $k_\TM$, such that $m^*(k_\TM) = \mG$
without changing the optimal (interpolating) solution.
This is shown schematically in Figure~\ref{fig:adaptive_vs_original}.
We see that for small mini-batch size convergence of these two kernels $k$ and $k_\TM$ is similar. However, values of $m>m^*(k)$ do not help the convergence of the original kernel $k$, while convergence of $k_\TM$ keep improving up to $m=\mG$, where the resource utilization is saturated.
For empirical results on real datasets, parallel to the schematic shown above, see Figure~\ref{fig:time-to-converge} in Section~\ref{sec:expr}.

We construct  and implement these kernels (see \href{https://github.com/EigenPro/EigenPro2}{github.com/EigenPro/EigenPro2} for the code), 
and show how to analytically choose  parameters, including the batch size and the step size. 
As a secondary contribution of this work we develop an improved version of EigenPro~\cite{ma2017diving} significantly reducing the memory requirements and making the computational overhead 
over the standard SGD negligible.
\newline

\noindent{\bf Comparison to related work.} 
In recent years there has been significant  progress on scaling and accelerating kernel methods including~\cite{takac2013mini,huang2014kernel,lu2014scale,tu2016large,avron2016faster,may2017kernel}. Most of these methods are able to scale to large data sets by utilizing major computational resources such as supercomputers or multiple (sometimes hundreds or thousands) AWS vCPU's\footnote{See \url{http://aws.amazon.com/ec2} for details.}.
Two recent methods which allow for high efficiency kernel training with a single CPU or GPU is EigenPro~\cite{ma2017diving} (used a as basis for the adaptive kernels in this paper) and FALKON~\cite{rudi2017falkon}.  The method developed in this paper is significantly faster than either of them, while achieving similar or better test set accuracy. Additionally, it is  easier to use as much of the parameter selection is done automatically. 

Mini-batch SGD (used in our algorithm) has been the dominant technique in training deep models.
There has been significant empirical evidence~\cite{krizhevsky2014one, you2017scaling, smith2017don} showing that
linearly scaling the step size with the mini-batch size up to a certain value leads to improved convergence. This phenomenon has been utilized to scale deep learning in distributed systems by adopting large mini-batch sizes~\cite{goyal2017accurate}.

The advantage of our setting is that  the optimal batch and step sizes can be analyzed and expressed analytically. Moreover, these formulas contain variables which can be explicitly computed and directly used for parameter selection in our algorithms.
Going beyond batch size and step size selection, the theoretical interpolation framework allows us to construct new adaptive kernels, such that the mini-batch size required for optimal convergence matches the capacity of the computational resource.
\newline

The paper is structured as follows:
In Section~\ref{sec:main}, we present our main algorithm to learn a kernel to fully utilize a given computational resource.
In Section~\ref{sec:epro-ne}, we present an improved version of EigenPro iteration used by the main algorithm. 
We then provide comparisons to  state-of-the-art kernel methods on several large datasets in Section~\ref{sec:expr}. We further discuss exploratory machine learning in the context of our method.

\section{Setup}

\label{sec:setup}

We start by briefly discussing the basic setting and kernel methods used in this paper.

\noindent {\bf Kernel interpolation.} 
We are given $n$ labeled training points
$(\bx_1, y_1), \ldots, (\bx_n, y_n) \in \mathbb{R}^d \times \mathbb{R}$. 
We consider a Reproducing Kernel Hilbert Space (RKHS) $\HH$~\cite{aronszajn1950theory}
corresponding  to a positive definite kernel function $\rmk: \mathbb{R}^d \times \mathbb{R}^d \rightarrow \mathbb{R}$.
There is   a unique (minimum norm) interpolated solution in $\HH$of the form
\begin{align*}
& f^*(\cdot) = \sum_{i=1}^n \alpha^*_i \rmk(\bx_i,\cdot),\,\\
& \mathrm{where} ~~ (\alpha^*_1,\ldots, \alpha^*_n)^T = \KM^{-1} (y_1,\ldots, y_n)^T    
\end{align*}

Here $\KM$ denotes an $n \times n$ kernel matrix, $\KM_{ij} = \rmk(\bx_i, \bx_j)$. It is easy to check that $\forall_i f^*(x_i)=y_i$. 

\begin{remark}[Square loss]
While the interpolated solution $f^*$ in $\HH$ does not depend on any loss function,
it is the unique minimizer in $\HH$ for the empirical square loss
$L(f) \defeq \frac{1}{n} \sum_{i=1}^n (f(\bx_i) - y_i)^2$.
\end{remark}



\noindent {\bf Gradient descent.} It can be shown that gradient descent iteration for the empirical squared loss in RKHS $\HH$ is given by 
\begin{equation}\label{eq:gd}
f \leftarrow
f - \eta \cdot \frac{2}{n}  \sum_{i=1}^n (f(\bx_i) - y_i) \rmk(\bx_i, \cdot)
\end{equation}


\noindent {\bf Mini-batch SGD.} Instead of calculating the gradient with $n$ training points, each SGD iteration updates the solution $f$ using $m$ subsamples $(\tx{1}, \ty{1}), \ldots, (\tx{m}, \ty{m})$,
\begin{equation}\label{eq:sgd}
f \leftarrow f - \eta \cdot \frac{2}{m}
\left\lbrace \sum_{i=1}^m (f(\tx{i}) - \ty{i}) \rmk(\tx{i}, \cdot) \right\rbrace
\end{equation}
It is equivalent to randomized coordinate descent~\cite{leventhal2010randomized} for $\KM \balpha = \by$ on $m$ coordinates of $\balpha$,
\begin{equation}\label{eq:sgd-c}
\alpha_{t_i} \leftarrow \alpha_{t_i} - \eta \cdot \frac{2}{m}
\left\lbrace f(\tx{i}) - \ty{i}
\right\rbrace
~~\text{for}~~
i = 1, \ldots, m
\end{equation}

\noindent {\bf Critical mini-batch size as effective parallelism.}
Theorem 4 in~\cite{ma2017interpolation} shows that for mini-batch iteration (\ref{eq:sgd}) with kernel $\rmk$ there is a data-dependent batch size
$m^*(\rmk)$ such that
\begin{itemize}
\item Convergence per iteration improves linearly with increasing  batch size$m$ for $m \leq m^*(\rmk)$ (using optimal constant step size).
\item Training with any batch size $m > m^*(\rmk)$ leads to the same convergence per iteration as training with $m^*(\rmk)$
up to a small constant factor.
\end{itemize}
We can calculate $m^*(\rmk)$ explicitly using kernel matrix $K$ (depending on the data),
$$
m^*(k) = \frac{\beta(K)}{\lambda_1(K)}
~~\text{where}~~ \beta(K) \defeq \max_{i=1,\ldots,n} k(\bx_i, \bx_i)
$$
For any shift invariant kernel $k$, after normalization, we have
$\beta(K) = \max_{i=1}^n k(\bx_i, \bx_i) \equiv 1 $.

\noindent {\bf EigenPro iteration~\cite{ma2017diving}.}
To achieve faster convergence,
EigenPro iteration performs spectral modification on the kernel operator $\rmH(f) \defeq \frac{2}{n} \sum_{i=1}^{n} \iprod{\rmk(\bx_i, \cdot)}{f}_\HH \rmk(\bx_i, \cdot)$ using operator, 
\begin{equation}\label{eq:ep-p}
\rmP(f) \defeq f - \sum_{i=1}^\epp
(1 - \frac{\lambda_{\epp}}{\lambda_i})
\iprod{\rme_i}{f}_\HH \rme_i
\end{equation}
where $\lambda_1 \geq \cdots \geq \lambda_n$ are ordered eigenvalues
of $\rmH$ and $e_i$ is its eigenfunction corresponding to $\lambda_i$.
The iteration uses $\rmP$ to rescale a (stochastic) gradient in $\HH$,
\begin{equation}\label{eq:ep-sgd}
f \leftarrow f - \eta \cdot \rmP
\left\lbrace \frac{2}{m} \sum_{i=1}^m (f(\tx{i}) - \ty{i}) \rmk(\tx{i}, \cdot) \right\rbrace
\end{equation}





\begin{remark}[Data adaptive kernel for fast optimization]
EigenPro iteration for target function $y$ and kernel $\rmk$ is equivalent to Richardson iteration / randomized (block) coordinate descent for linear system $\KM_{\rmP} \balpha = \by_{\rmP} \defeq (\rmP f^* (\bx_1), \ldots, \rmP f^* (\bx_n))^T$.
Here $\KM_{\rmP}$ is the kernel matrix corresponding to a data-dependent kernel $\rmk_{\rmP}$. When $n \rightarrow \infty$, it has the following expansion according to Mercer's theorem,
\begin{equation}\label{eq:Kp}
\rmk_{\rmP}(\bx, \bz) =
\sum_{i=1}^\epp \lambda_\epp \rme_i(\bx) \rme_i(\bz)
+ \sum_{i=\epp+1}^{\infty} \lambda_i \rme_i(\bx) \rme_i(\bz)
\end{equation}
For $n < \infty$, it is a modification of the original kernel $\rmk$,
\begin{equation*}
\begin{split}
\rmk_{\rmP}(\bx, \bz) & =
\rmP\{\rmk(\bx, \cdot)\} (\bz) \\
&\approx \rmk(\bx, \bz)
- \sum_{i=1}^\epp (\lambda_i - \lambda_\epp) \rme_i(\bx) \rme_i(\bz)
\end{split}
\end{equation*}
\end{remark}

\begin{remark}[Preconditioned linear system / gradient descent]
$\KM_{\rmP} \balpha = \by_{\rmP}$ is equivalent to the preconditioned linear system
$P \KM \balpha = P \by$ where $P$ is a left matrix preconditioner related to $\rmP$.
Accordingly, $\rmP$ is the operator preconditioner for preconditioned (stochastic) gradient descent (\ref{eq:ep-sgd}).
\end{remark}

\noindent {\bf Abstraction for parallel computational resources.}
To construct a resource adaptive kernel, we consider the following abstraction for given computational resource $\TM$,
\begin{itemize}
\item $C_{\TM}$: Parallel capacity of $\TM$, i.e., the number of parallel operations that is required to  fully utilize the computing capacity of $\TM$.
\item $S_{\TM}$: Internal resource memory of $\TM$.
\end{itemize}
To fully utilize $\TM$, one SGD~/~EigenPro iteration must execute at least $C_{\TM}$ 
operations using less than $S_{\TM}$ memory.
In this paper, we primarily adapt kernel to GPU devices.
For a GPU $\TM$, $S_{\TM}$ equals the size of its dedicated memory and $C_{\TM}$ is proportional to the number of the computing cores (e.g., 3840 CUDA cores in Titan Xp).
Note for computational resources like cluster and supercomputer, we need to take into account additional factors such as network bandwidth.

\section{Main Algorithm}\label{sec:main}

Our main algorithm aims to reduce the training time  by constructing a data/resource adaptive kernel for any given kernel function $\rmk$ to fully utilize a computational resource $\TM$.
Its detailed workflow is presented on the right. 
Specifically, we use the following steps: 
\begin{itemize}
\item[\bf Step 1.] Calculate  the resource-dependent  mini-batch size $\mG$ to fully utilize resource $\TM$.
\item[\bf Step 2.] Identify the parameters and construct a new kernel ${k_{\TM}}$ such that $m^*(k_{\TM}) = \mG$ .
\item[\bf Step 3.] Select optimal step size  and train using improved EigenPro (see Section~\ref{sec:epro-ne}).
\end{itemize}

\begin{wrapfigure}{r}{0.25\textwidth}
\vspace{-4mm}
\begin{minipage}{0.23\textwidth}
  \includegraphics[width=\textwidth]{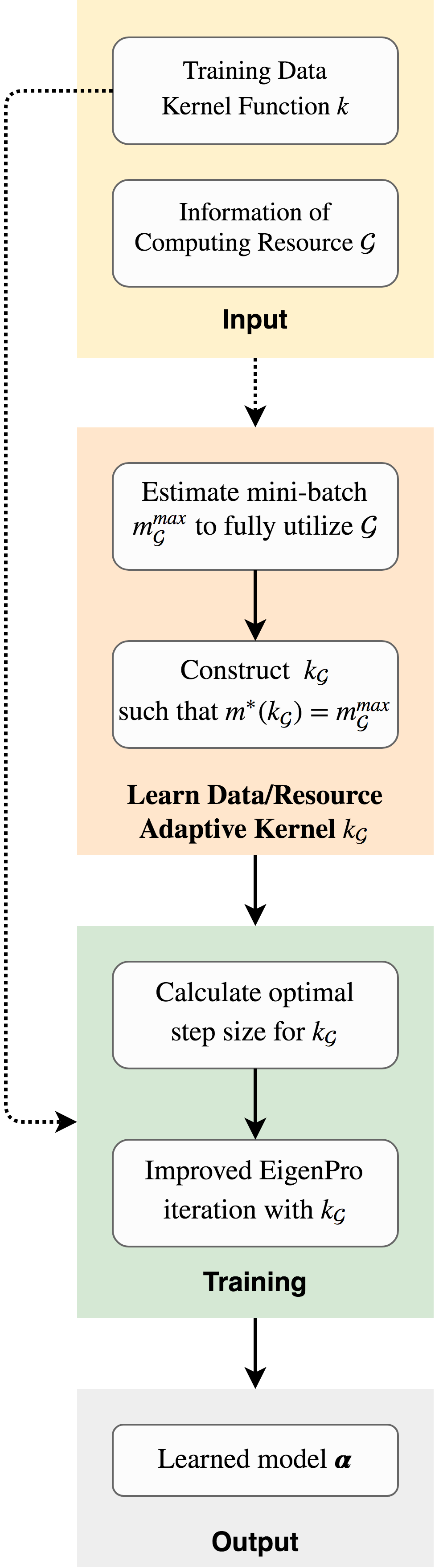}
	\caption*{~~~~~~~EigenPro 2.0}
\end{minipage}
\vspace{-14mm}
\end{wrapfigure}

Note that due to  properties of EigenPro iteration, training with this adaptive kernel converges to the same  solution as the original kernel.

To calculate $\mG$ for 100\% resource utilization, we first estimate the operation parallelism and memory usage of one EigenPro iteration.
The improved version of EigenPro iteration (introduced in Section~\ref{sec:epro-ne})  makes computation and memory overhead over the standard SGD negligible (see Table~\ref{tbl:overhead}). Thus
we  assume that EigenPro has the same complexity as the standard  SGD per iteration.

\noindent {\bf Cost of one EigenPro iteration with batch size $m$.}
We consider training data 
 $(\bx_i, \by_i) \in \mathbb{R}^d \times \mathbb{R}^l, i=1,\ldots, n$.
Here each feature vector $\bx$ is $d$ dimensional, and each label $\by$ is $l$ dimensional.
\begin{itemize}
\item {\bf Computational cost.}
It takes $(d + l) \cdot m \cdot n$  operations to perform one SGD iteration on $m$ points as in Iteration~(\ref{eq:sgd}). These computations reduce to matrix multiplication and can be done in parallel. 
\item {\bf Space usage.}
It takes $d \cdot n$ memory to store the training data (as kernel centers) and  $l \cdot n$ memory to maintain the model weight.
Additionally we need to store   a $m \cdot n$ kernel matrix for the prediction on the mini-batch.
In total, we need $(d + l + m) \cdot n$ memory.
\end{itemize}

We can now calculate $\mG$ for the parallel computational resource $\TM$ with parameters ${C_{\TM}},{S_{\TM}}$ and introduced in Section~\ref{sec:setup}.

\noindent {\bf Step 1: Determining batch size $\mG$ for 100\% resource utilization.} We first define two mini-batch notations:
\begin{itemize}
\item $m_{C_{\TM}}$: batch size for fully utilizing parallelism in $\TM$ such that $(d + l) \cdot m_{C_{\TM}} \cdot n \approx C_{\TM}$.
\item $m_{S_{\TM}}$: batch size for maximum memory usage of $\TM$ such that
$(d + l + m_{S_{\TM}}) \cdot n \approx S_{\TM}$.
\end{itemize}
To best utilize $\TM$ without exceeding its memory, we set $\mG = \min \{ m_{C_{\TM}}, m_{S_{\TM}} \}$. Note that in practice, it is more important to fully utilize the memory so that $\mG \lesssim m_{S_{\TM}}$.\\


\noindent {\bf Step 2: Learning the kernel $k_{\TM}$ given $\mG$.}
Next, we show how to construct $\rmk_{\TM}=\rmk_{\rmP_\epp}$ using EigenPro iteration such that
$m^*(k_{\TM}) = \mG$.
The corresponding $\epp$ is defined as
\begin{equation}\label{eq:q}
\epp \defeq \max ~\{i\in \mathbb{N},~\text{s.t.}~~ m^*(k_{\rmP_i}) \le \mG \}
\end{equation}
To compute $\epp$ recall that 
$m^*(k_{\rmP_\epp}) = \frac{\beta(K_{\rmP_\epp})}{\lambda_1(K_{\rmP_\epp})}$,
where $K_{\rmP_\epp}$ is the kernel matrix corresponding to the kernel function  $k_{\rmP_\epp}$.
Using the definition of
$\rmP_\epp$ and $\beta$ in Section~\ref{sec:setup}, we have 
\begin{equation*}
\begin{split}
\lambda_1(K_{\rmP_\epp}) &= \lambda_\epp(K)\\
\beta(K_{\rmP_\epp}) &\approx \max_{i=1,\ldots,n}{\rmk_{\rmP_\epp}(\bx_i, \bx_i)} \\
& = \max_{i=1,\ldots,n} \{ \rmk(\bx_i, \bx_i) - 
\sum_{j=1}^\epp (\lambda_j - \lambda_\epp) \rme_i(\bx_i)^2\}
\end{split}
\end{equation*}
In practice, $\beta(K_{\rmP_\epp})$ can be accurately estimated using the maximum of $\rmk_{\rmP_\epp}(\bx, \bx)$ on a small number of subsamples.
Similarly, we can estimate $\lambda_\epp(K)$ on a subsample kernel matrix. Knowing the approximate top eigenvalues of  $K$, allows us to efficiently compute $m^*(\rmk_{\rmP_p})$ for each $p$,  
thus allowing to choose $\epp$ from (\ref{eq:q}).

\noindent {\bf Step 3: Training with adaptive kernel $\rmk_{\TM}=\rmk_{\rmP_\epp}$.}
We use the learned kernel $k_\TM$ with improved EigenPro (Section~\ref{sec:epro-ne}).
Its optimization parameters (batch and step size) are calculated as follows:
$$
m = \mG, \eta = \frac{\mG}{\beta(K_{\TM})}
$$

\noindent {\bf Claim (Acceleration).} 
Using the adaptive kernel $\rmk_\TM$ decreases the resource time required for training (assuming an idealized model of the GPU and  workload)  over the original kernel $k$ by a factor of 
$$
\text{acceleration of}~ k_\TM ~\text{over}~ k = \frac{\beta(K)}{\beta(K_\TM)} \cdot \frac{\mG}{m^*(k)}
$$  
See the Appendix~\ref{sec:app-analysis} for the derivation and a discussion. \\
We note that empirically, ${\beta(K_\TM)} \approx {\beta(K)}$, while $ \frac{\mG}{m^*(k)}$ is between  $50$ and $500$, which is in line with the acceleration observed in practice.

\begin{remark}[Choice of $q$] Note that it is not important to select  $q$ exactly, according to Eq.~\ref{eq:q}. 
In fact, choosing $\rmk_{\rmP_{p}}$ for any $p > \epp$ allows for the same acceleration as  $\rmk_{\rmP_{\epp}}$ as long as the mini-batch size is chosen to be $\mG$ and the step size is chosen accordingly. Thus, we can choose any value $p > \epp$ for our adaptive kernel $k_{\rmP_{p}}$.
However, choosing $p$ larger than $q$ incurs an additional computation cost as $p$ eigenvalues and eigenvectors of $K$ need to be approximated accurately. In particular, larger subsample size $s$ (see Section~\ref{sec:epro-ne} may be needed for approximating eigenvectors. 
\end{remark}


\section{Improved EigenPro Iteration using Nyström Extension}
\label{sec:epro-ne}

In this section, we present an improvement for the EigenPro iteration originally proposed in~\cite{ma2017diving}.
We significantly reduce the memory overhead of EigenPro over standard SGD and nearly eliminate 
computational overhead per iteration. The improvement is based on an efficient representation of the preconditioner $\rmP_\epp$ using Nyström extension. 
 
We start by recalling the EigenPro iteration in RKHS
and its preconditioner constructed by the top-$\epp$ eigensystem $\lambda_i, e_i$ of the kernel operator $\rmH$:
\begin{equation*}
\begin{split}
& f \leftarrow f - \eta \cdot \rmP_\epp
\left\lbrace \frac{2}{m} \sum_{i=1}^m (f(\tx{i}) - \ty{i}) \rmk(\tx{i}, \cdot) \right\rbrace\\
&\text{where}~~
\rmP_\epp(f) = f - \sum_{i=1}^\epp
(1 - \frac{\lambda_{\epp}}{\lambda_i})
\iprod{\rme_i}{f}_\HH \rme_i
\end{split}
\end{equation*}
The key to construct the above iteration is to obtain an accurate and computationally efficient approximation of $\lambda_i, e_i$ such that
$\rmH e_i \approx \lambda_i e_i$.
The original EigenPro iteration learns an approximate $e_i$ of the form $\sum_{j=1}^n w_j k(\bx_j, \cdot)$.
In contrast, our improved version of Eigenpro uses only a small number of subsamples
$\rx{1}, \ldots,\rx{s}$
to learn an $e_i$ of the form
$\sum_{j=1}^s w_j k(\rx{j}, \cdot)$.
This compact representation ($s$ versus $n$) nearly eliminates per-iteration overhead of EigenPro over SGD. Importantly, there is no associated accuracy reduction as this is the same subset used in the original EigenPro to approximate $\rmP_\epp$.

\begin{algorithm}[H]
\caption{Improved EigenPro iteration\\
(double coordinate block descent)}
\label{alg:epro}
\begin{algorithmic}

\STATE{\bf Input:} Kernel function $k(\bx, \bz)$,
EigenPro parameter $\epp$,
mini-batch size $m$, step size $\eta$,
size of fixed coordinate block $s$
\STATE{}
\vspace{-1.5mm}

\STATE{\bf initialize} model parameter $\balpha = (\alpha_1, \ldots, \alpha_n)^T \leftarrow 0$

\STATE{\bf subsample} $s$ coordinate indices $r_1, \ldots, r_s \in \left\{1, \ldots, n\right\}$ for constructing $\rmP_\epp$,
which form fixed coordinate block 
$\balpha_r \defeq (\alpha_{r_1}, \ldots, \alpha_{r_s})^T$

\STATE{\bf compute} top-$\epp$ eigenvalues
$\Sigma \defeq \text{diag}(\sigma_1, \ldots, \sigma_\epp)$ and corresponding eigenvectors
$V \defeq (\be_1, \ldots, \be_\epp)$ of
subsample kernel matrix
$K_s = [\rmk(\rx{i}, \rx{j})]_{i, j = 1}^s$

\STATE{}
\vspace{-1.5mm}
\FOR{$t = 1, \ldots$}
\STATE 1. sample a mini-batch $(\tx{1}, \ty{1}), \ldots, (\tx{m}, \ty{m})$

\STATE 2. calculate predictions on the mini-batch\\
\vspace{-6mm}
$$
f(\bx_{t_j}) = \sum_{i=1}^n \alpha_i \rmk(\bx_i,\bx_{t_j})
~~\textnormal{for}~~ j = 1, \ldots, m
$$
\vspace{-4mm}

\STATE 3. update {\bf sampled coordinate block} corresponding to the mini-batch
$\balpha_t \defeq (\alpha_{t_1}, \ldots, \alpha_{t_m})$,
\vspace{-2mm}
$$
\balpha_t \leftarrow \balpha_t - \eta \cdot
\frac{2}{m} (f(\tx{1}) - \ty{1}, \ldots, f(\tx{m}) - \ty{m})^T
$$
\vspace{-4mm}

\STATE 4. evaluate the following feature map $\phi(\cdot)$ on the mini-batch features $\bx_{t_1}, \ldots, \bx_{t_m}$:\\
\vspace{-2mm}
$$
\phi(\bx) \defeq (\rmk(\rx{1}, \bx), \ldots, \rmk(\rx{s}, \bx))^T
$$

\STATE 5. update {\bf fixed coordinate block}
$\balpha_r$ to apply $\rmP_\epp$,
\vspace{-2mm}
\begin{equation*}
\begin{split}
\balpha_r \leftarrow & \balpha_r + \eta \cdot \frac{2}{m}
\sum_{i=1}^m
(f(\tx{i}) - \ty{i}) \cdot
VDV^T \phi(\tx{i}) \\
&\textnormal{where}~~ D \defeq (1 - \sigma_\epp \cdot \Sigma^{-1})\Sigma^{-1}
\end{split}
\end{equation*}
\vspace{-4mm}
\ENDFOR

\end{algorithmic}
\end{algorithm}

Next, we show how to approximate $\lambda_i, e_i$.
We first consider a related linear system for subsamples $\rx{1}, \ldots, \rx{s} \in \mathbb{R}^d$:
$K_s \be_i = \sigma_i \be_i$
where $K_s \defeq [\rmk(\rx{i}, \rx{j})]_{i, j = 1}^s$ is a subsample kernel matrix and $\sigma_i, \be_i$ is its eigenvalue/eigenvector.
This rank-$s$ linear system is in fact a discretization of
$\rmH e_i = \lambda_i e_i$ in the RKHS.

The two eigensystems, $\sigma_i, \be_i$ and $\lambda_i, e_i$ are connected  through {\it Nyström extension}.
Specifically, the Nyström extension of $e_i$ on subsamples $\rx{1}, \ldots, \rx{s}$ approximates $e_i$ as follows:
$$
e_i(\cdot)
\approx \frac{1}{\sigma_i}
\sum_{j=1}^{s} e_i(\rx{j}) \rmk(\rx{j}, \cdot)
$$
Evaluating both side on $\rx{1}, \ldots, \rx{s}$, we have
\begin{equation*}
\lambda_i \approx \frac{\sigma_i}{s},
e_i(\cdot) \approx
\frac{1}{\sqrt{\sigma_i}} \be_i^T \phi(\cdot)
\end{equation*}
where $\phi(\cdot) \defeq (\rmk(\rx{1}, \cdot), \ldots, \rmk(\rx{s}, \cdot))^T$ is a kernel feature map.
Thus we approximate the top-$\epp$ eigensystem of $\rmH$ using the top-$\epp$ eigensystem of $K_s$.
These (low-rank) approximations further allow us to apply $\rmP_\epp$ for efficient EigenPro iteration on mini-batch $(\tx{1}, \ty{1}) \ldots, (\tx{m}, \ty{m})$,
\begin{equation}\label{eq:nep-sgd}
\begin{split}
f &\leftarrow 
f - \eta \cdot \frac{2}{m} \sum_{i=1}^m (f(\tx{i}) - \ty{i}) \rmk(\tx{i}, \cdot)\\
&+ \eta \cdot
\frac{2}{m} \sum_{i=1}^m (f(\tx{i}) - \ty{i}) \cdot
\phi(\tx{i})^T V D V^T \phi(\cdot)\\
& \textnormal{where}~~
D \defeq \Sigma^{-1}(1 - \sigma_\epp \cdot \Sigma^{-1})
\end{split}
\end{equation}
where
$\Sigma \defeq \text{diag}(\sigma_1, \cdots, \sigma_\epp)$ and
$V \defeq (\be_1, \cdots, \be_\epp)$ are top-$\epp$ eigensystem of $K_s$.\\

Recalling that $f = \sum_{i=1}^n \alpha_i \rmk(\bx_i,\cdot)$, the above iteration can be executed by updating two coordinate blocks of the parameter vector $\balpha$ as in Algorithm~\ref{alg:epro}.

\noindent {\bf Computation/memory per  iteration.}
In Algorithm~\ref{alg:epro}, the  cost of each iteration relates to updating two coordinate blocks. 
Notably, Steps 2-3 is exactly the standard SGD. Thus the overhead of our method comes from Steps 4-5. 
We compare our improved EigenPro to the original EigenPro and to standard SGD in Table~\ref{tbl:overhead}. We see that the overhead of original EigenPro (in bold) scales with the data size $n$. In contrast, improved EigenPro  depends only on the fixed coordinate block size $s$ which is independent of $n$.  
Hence, when $n$ becomes large, the overhead of our iteration becomes negligible (both in computation and memory)  compared to the cost of SGD.
\begin{table}[!ht]
\centering
\begin{adjustbox}{center}
\resizebox{.5\textwidth}{!}{
\begin{tabular}{|c||c||c|}
\hline
 & Computation & Memory \\ \hline
Improved EigenPro & $\pmb{s \cdot mq} + n \cdot m (d + l)$ & $\pmb{s \cdot q} + n \cdot (m + d + l)$ \\ \hline
Original EigenPro & $\pmb{n \cdot mq} + n \cdot m (d + l)$ & $\pmb{n \cdot q}+ n \cdot (m + d + l)$ \\ \hline
SGD & $n \cdot m (d + l)$ & $n \cdot (m + d + l)$ \\ \hline
\end{tabular}
}
\end{adjustbox}
\caption{Overhead over SGD is {\bf bolded}. $n$: training data size, $m$: batch size, $d$: feature dim., $s$: fixed coordinate block size, $q$: EigenPro parameter,
$l$: number of labels.}
\label{tbl:overhead}
\end{table}

To give a realistic example, for many of our experiments $n=10^6$, while $s$ is chosen to be $10^4$. We typically have $d,m$ of the same order of magnitude $~10^3$, while $\epp$ and $l$ around $10^2$. This results in overhead of EigenPro of less than $1\%$ over SGD for both computation and memory.\\

\begin{table*}[!ht]
\caption{Comparison of EigenPro 2.0 and state-of-the-art kernel methods}
\label{tbl:sota}
\begin{threeparttable}
\vspace{-1mm}
\begin{adjustbox}{center}
\resizebox{\textwidth}{!}{
\begin{tabular}{|c|c||c|c||c|c|c|}
\hline
\multirow{2}{*}{Dataset} & \multirow{2}{*}{Size} & \multicolumn{2}{c||}{\begin{tabular}[c]{@{}c@{}}EigenPro 2.0\\ (use 1 GTX Titan Xp)\end{tabular}} & \multicolumn{3}{c|}{Results of Other Methods} \\ \cline{3-7} 
 &  & error & GPU time & resource time & error & reference \\ \hline \hline
\multirow{3}{*}{MNIST} & \multirow{3}{*}{$6.7 \times 10^6$} & \multirow{3}{*}{0.72\%} & \multirow{3}{*}{\textbf{19 m}} & \begin{tabular}[c]{@{}c@{}}4.8 h on\\ 1 GTX Titan X\end{tabular} & 0.70\% & EigenPro~\cite{ma2017diving} \\ \cline{5-7} 
 &  &  &  & \begin{tabular}[c]{@{}c@{}}1.1 h on\\ 1344 AWS vCPUs\end{tabular} & 0.72\% & PCG~\cite{avron2016faster} \\ \cline{5-7} 
 &  &  &  & \begin{tabular}[c]{@{}c@{}}less than 37.5 hours\\ on 1 Tesla K20m\end{tabular} & 0.85\% & \cite{lu2014scale} \\ \hline \hline
\multirow{2}{*}{ImageNet$^\dagger$} & \multirow{2}{*}{$1.3 \times 10^6$} & \multirow{2}{*}{20.6\%} & \multirow{2}{*}{\textbf{40 m}} & - & 19.9\% & Inception-ResNet-v2~\cite{szegedy2017inception} \\ \cline{5-7} 
 &  &  &  & \begin{tabular}[c]{@{}c@{}}4 h on \\ 1 Tesla K40c\end{tabular} & 20.7\% & FALKON~\cite{rudi2017falkon} \\ \hline\hline
\multirow{6}{*}{TIMIT$^\ddagger$} & \multirow{6}{*}{\begin{tabular}[c]{@{}c@{}}$1.1 \cdot 10^6$\\ / $2 \cdot 10^6$\end{tabular}} & \multirow{6}{*}{\begin{tabular}[c]{@{}c@{}}{}\\ {}\\ 31.7\%\\ \\ \\ \\ 32.1\%\end{tabular}} & \multirow{6}{*}{\begin{tabular}[c]{@{}c@{}}{}\\ {}\\ 24 m\\ (3 epochs)\\ \\ \\ \textbf{8 m}\\ (1 epoch)\end{tabular}} & \begin{tabular}[c]{@{}c@{}}3.2 h on\\ 1 GTX Titan X\end{tabular} & 31.7\% & EigenPro~\cite{ma2017diving} \\ \cline{5-7} 
 &  &  &  & \begin{tabular}[c]{@{}c@{}}1.5 h on\\ 1 Tesla K40c\end{tabular} & 32.3\% & FALKON~\cite{rudi2017falkon} \\ \cline{5-7} 
 &  &  &  & \begin{tabular}[c]{@{}c@{}}512 IBM\\ Blue Gene/Q cores\end{tabular} & 33.5\% & Ensemble~\cite{huang2014kernel} \\ \cline{5-7} 
 &  &  &  & \begin{tabular}[c]{@{}c@{}}7.5 h on \\ 1024 AWS vCPUs\end{tabular} & 33.5\% & BCD~\cite{tu2016large} \\ \cline{5-7} 
 &  &  &  & \begin{tabular}[c]{@{}c@{}}multiple AWS\\ g2.2xlarge instances\end{tabular} & 32.4\% & DNN~\cite{may2017kernel} \\ \cline{5-7} 
 &  &  &  & \begin{tabular}[c]{@{}c@{}}multiple AWS \\ g2.2xlarge instances\end{tabular} & 30.9\% & \begin{tabular}[c]{@{}c@{}}SparseKernel~\cite{may2017kernel}\\ (use learned features)\end{tabular} \\ \hline\hline
\multirow{3}{*}{SUSY} & \multirow{3}{*}{$4 \cdot 10^6$} & \multirow{3}{*}{19.7\%} & \multirow{3}{*}{\textbf{58 s}} & \begin{tabular}[c]{@{}c@{}}6 m on\\ 1 GTX Titan X\end{tabular} & 19.8\% & EigenPro~\cite{ma2017diving} \\ \cline{5-7} 
 &  &  &  & \begin{tabular}[c]{@{}c@{}}4 m on\\ 1 Tesla K40c\end{tabular} & 19.6\% & FALKON~\cite{rudi2017falkon} \\ \cline{5-7} 
 &  &  &  & \begin{tabular}[c]{@{}c@{}}36 m on\\ IBM POWER8\end{tabular} & $\approx 20\%$ & Hierarchical~\cite{chen2016hierarchically} \\ \hline
\end{tabular}
}
\end{adjustbox}
\vspace{2mm}
\begin{tablenotes}[flushleft]
\begin{minipage}{17cm}
\footnotesize
\item $\dagger$ Our method uses the convolutional features from Inception-ResNet-v2 and Falkon uses the convolutional features from Inception-v4. Both neural network models are presented in~\cite{szegedy2017inception} and show nearly identical performance.
\item $\ddagger$ There are two sampling rates for TIMIT, which result in two training sets of different sizes.
\end{minipage}
\end{tablenotes}
\end{threeparttable}
\end{table*}

\section{Experimental Evaluation}\label{sec:expr}

\noindent {\bf Computing resource.} We run all experiments on a single workstation equipped with 128GB main memory, two Intel Xeon(R) E5-2620 processors, and one Nvidia GTX Titan Xp (Pascal) GPU.


\begin{figure*}[!ht]
  \centering
  \begin{minipage}[l]{0.47\textwidth}
    \includegraphics[width=\textwidth]{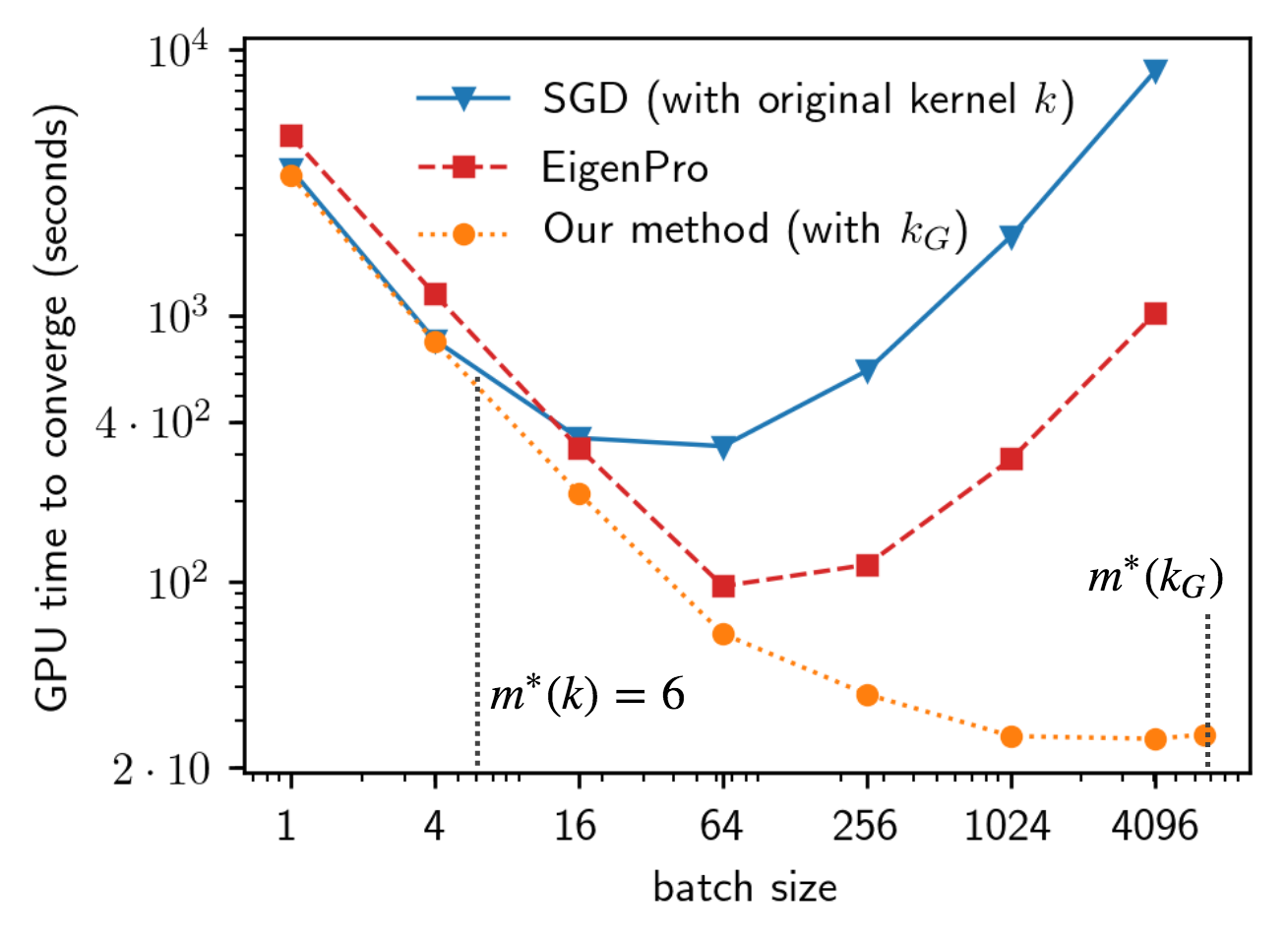}
    \vspace{-7mm}
    \subcaption{MNIST ($10^5$ subsamples), stop when train mse $ < 1\cdot 10^{-4}$}
  \end{minipage}
  \hfill
  \begin{minipage}[l]{0.47\textwidth}
    \includegraphics[width=\textwidth]{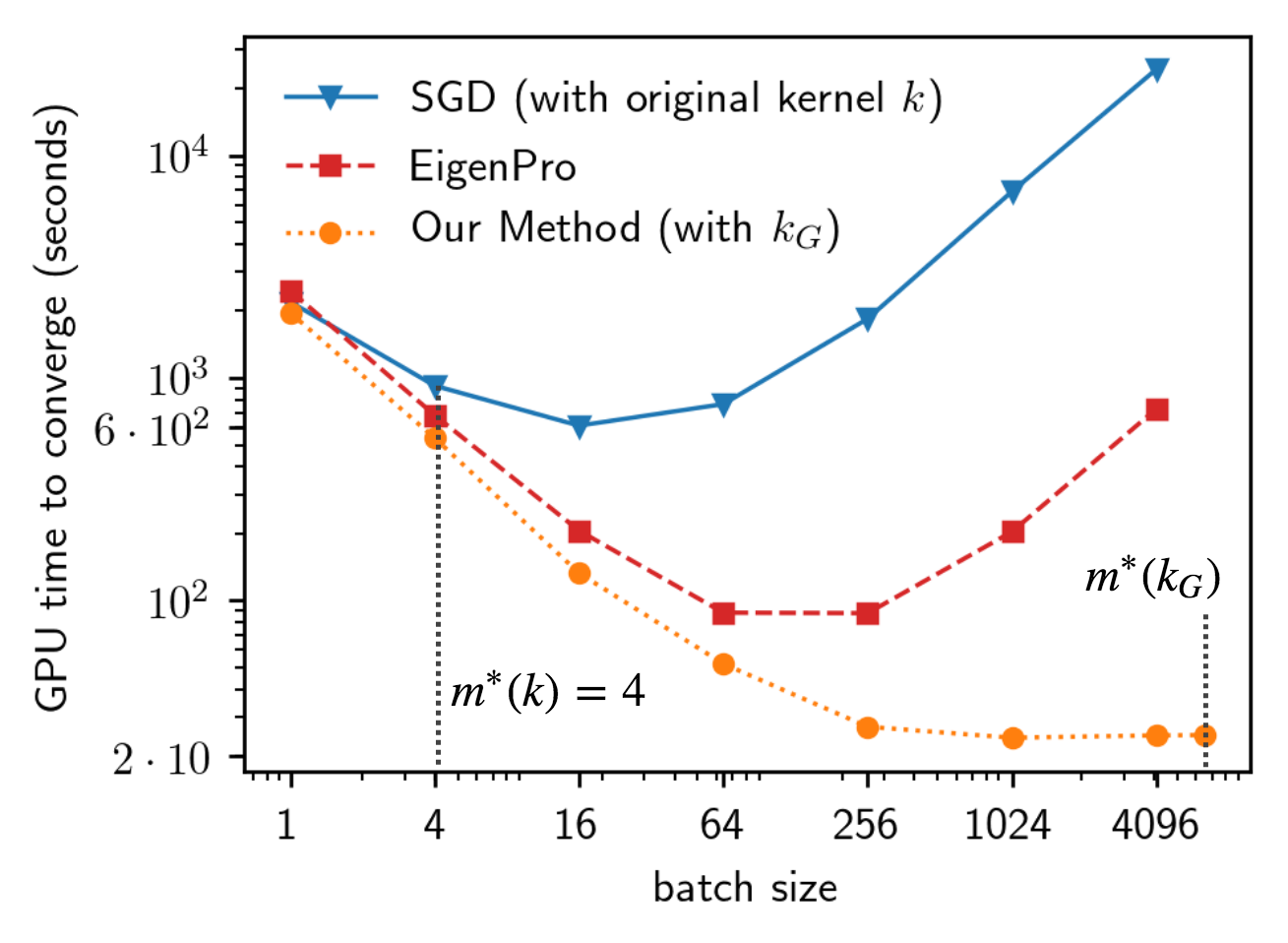}
    \vspace{-7mm}
    \subcaption{TIMIT ($10^5$ subsamples), stop when train mse $ < 2\cdot 10^{-4}$}
    \label{fig:timit-t2c}
  \end{minipage}
  \vspace{-2mm}
  \caption{Time to converge with different batch sizes and optimal step sizes}
  \label{fig:time-to-converge}
\vspace{-4mm}
\end{figure*}
\noindent {\bf Datasets.}
We reduce multiclass labels to multiple binary labels.
For image datasets including MNIST~\cite{lecun1998gradient}, CIFAR-10~\cite{krizhevsky2009learning}, and SVHN~\cite{netzer2011reading}, color images are first transformed to grayscale images. We then rescale the range of each feature to
$[0, 1]$.
For ImageNet~\cite{deng2009imagenet}, we use the top 500 PCA components of some convolutional features extracted from Inception-ResNet-v2~\cite{szegedy2017inception}.
For TIMIT~\cite{garofolo1993darpa}, we normalize each feature by z-score. 

\noindent {\bf Choosing the size of the fixed coordinate block $s$}. We choose $s$ according to the size of the training data, $n$. When $n \leq 10^5$, we choose $s = 2 \cdot 10^3$; when $n > 10^5$, we choose $s = 1.2 \cdot 10^4$.

\subsection{Comparison to state-of-the-art kernel methods}
In Table~\ref{tbl:sota}, we compare our method to the state-of-the-art kernel methods on several large datasets.
For all datasets, our method is significantly faster  than other methods while still achieving better or similar results.
Moreover, our method uses only a single GPU while many state-of-the-art kernel methods use much less accessible computing resources.

Among all the compared methods, FALKON~\cite{rudi2017falkon} and EigenPro~\cite{ma2017diving} stand out for their competitive performance and fast training on a single GPU.
Notably, our method still achieves 5X-6X acceleration over FALKON and 5X-14X acceleration over EigenPro with mostly better accuracy.
Importantly, our method has the advantage of automatically inferring parameters for optimization. In contrast, parameters related to optimization for FALKON and EigenPro need to be selected by cross-validation.

\begin{figure*}[!ht]
  \centering
  \begin{minipage}[l]{0.47\textwidth}
    \includegraphics[width=\textwidth]{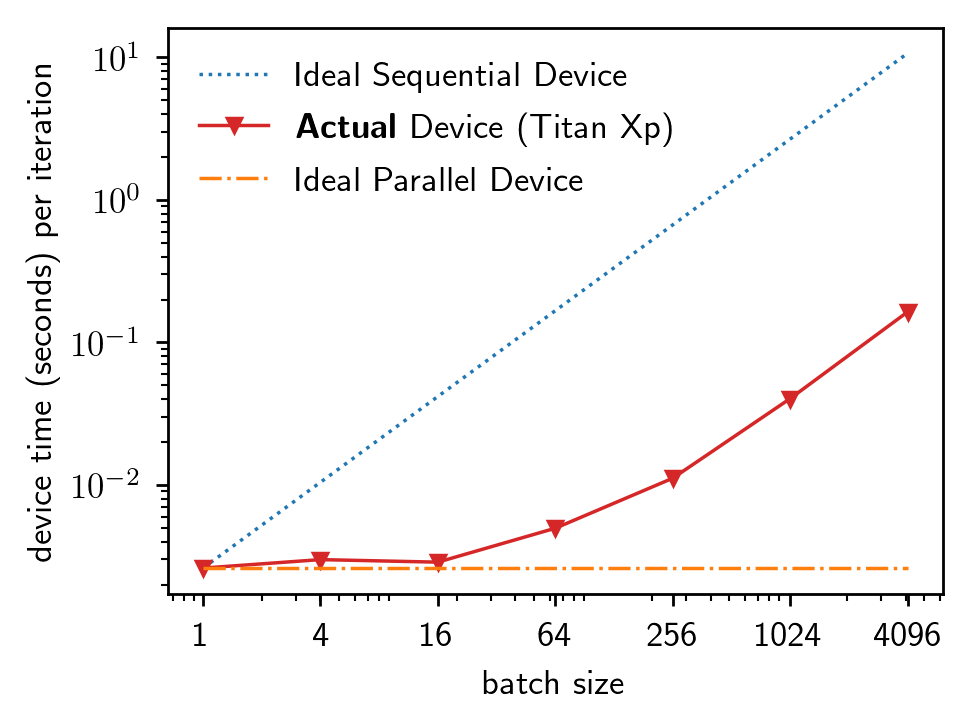}
    \vspace{-2mm}
    \subcaption{Time per training iteration of different batch sizes on actual and ideal devices (TIMIT,  $n = 10^5$, $d=440$ )\\}
    \label{fig:t/i}
  \end{minipage}
  \hfill
  \begin{minipage}[l]{0.47\textwidth}
    \includegraphics[width=\textwidth]{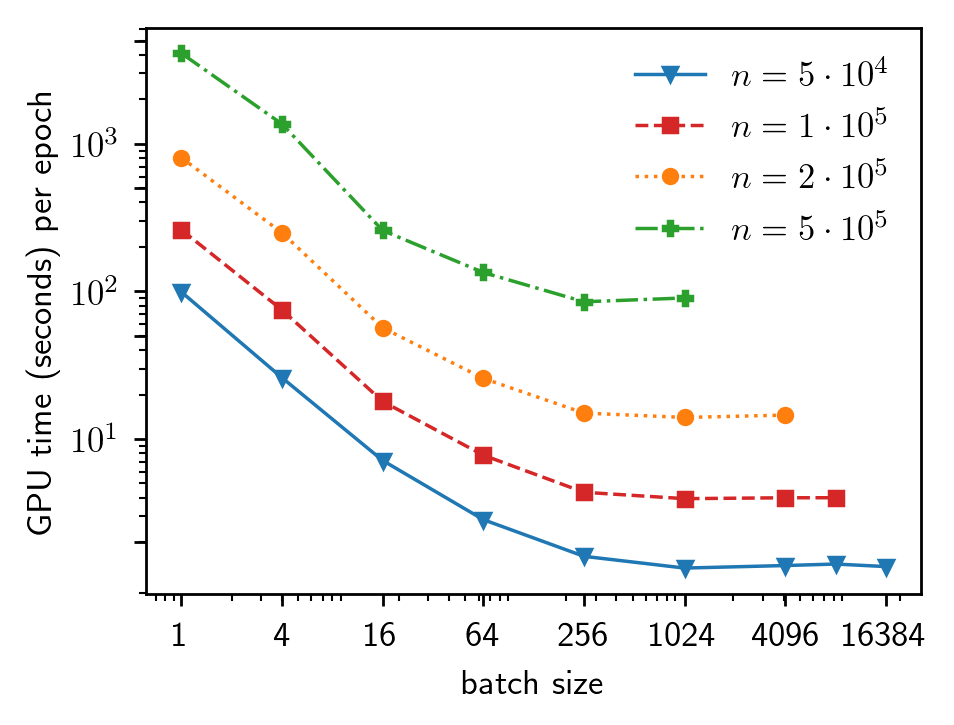}
    \vspace{-2mm}
    \subcaption{Time per training epoch on GPU with different sizes of train set ($n$, which is also the model size) and batch that fit into the GPU memory}
    \label{fig:t/e}
  \end{minipage}
  \caption{Time per iteration / epoch for training with different batch sizes}
\end{figure*}

\subsection{Convergence comparison to SGD and EigenPro}
In Figure~\ref{fig:time-to-converge}, we train three kernel machines with EigenPro~2.0, standard SGD and EigenPro~\cite{ma2017diving}  for various batch sizes. The step sizes for SGD and EigenPro are tuned  for best performance. The step size for EigenPro~2.0 is computed automatically according to Section~\ref{sec:main}.

Consistent with the schematic Figure~\ref{fig:adaptive_vs_original} in the introduction, the original kernel $k$ has a critical batch size $m^*(k)$ of size $4$ and $6$ respectively, which is too small to fully utilize the parallel computing capacity of the GPU device. In contrast, our adaptive kernel $k_G$ has a much larger critical batch size $m^*(k_G) \approx 6500$, which leads to maximum GPU utilization. We see that EigenPro~2.0 significantly outperforms original EigenPro due to better resource utilization and parameter selection, as well as lower overhead (see Table~\ref{tbl:overhead}).

\subsection{Batch size and GPU utilization}

The number of operation required for one iteration of SGD is linear in the batch size. Thus we expect that time required per iteration for  a pure sequential machine would scale linearly with batch size. On the other hand an ideal parallel device with no overhead requires the same amount of time to process any mini-batch.
In Figure~\ref{fig:t/i}, we show how the training time per iteration for actual GPU depends on the batch size. 
We see that for small batch sizes time per iteration is nearly constant, like that of an ideal parallel device, and start to  increase for larger batches.

Note that in addition to time per iteration we need to consider the overhead associated to each iteration. Larger batch sizes incur less overhead per epoch. This phenomenon is known in the systems literature as Amdahl's law~\cite{rodgers1985improvements}. 
In Figure~\ref{fig:t/e} we show GPU time  {\it per epoch}  for different model (training set) size ($n$).
We see consistent speed-ups by increasing mini-batch size across model sizes up to maximum GPU utilization.

\subsection{``Interactive'' 
training for exploratory machine learning}


\begin{table}[!h]
\centering
\begin{adjustbox}{center}
\resizebox{0.5\textwidth}{!}{
\begin{tabular}{|c|c|c||c|c|c|}
\hline
Dataset & Size & Feature & \begin{tabular}[c]{@{}c@{}}EigenPro\\ (GPU)\end{tabular} & \begin{tabular}[c]{@{}c@{}}ThunderSVM\\ (GPU)\end{tabular} & \begin{tabular}[c]{@{}c@{}}LibSVM\\ (CPU)\end{tabular} \\ \hline
TIMIT & $1 \cdot 10^5$ & 440 & \textbf{15 s} & 480 s & 1.6 h \\ \hline
SVHN & $7 \cdot 10^4$ & 1024 & \textbf{13 s} & 142 s & 3.8 h \\ \hline
MNIST & $6 \cdot 10^4$ & 784 & \textbf{6 s} & 31 s & 9 m \\ \hline
CIFAR-10 & $5 \cdot 10^4$ & 1024 & \textbf{8 s} & 121 s & 3.4 h \\ \hline
\end{tabular}
}
\end{adjustbox}
\caption{Comparing training time of kernel machines} 
\label{tbl:small}
\end{table}
Most practical tasks of machine learning require multiple training runs  for parameter and feature selection,  evaluating appropriateness of data or features to a given task, and various other exploratory purposes. 
While using hours, days or even months of machine time may be necessary to improve on the state of the art in large-scale certain problems, it is too time-consuming and expensive  for most  data analysis work. Thus, it is very desirable to train classifiers in close to real time.
One of the  advantages of our approach is the combination of its speed on small and medium datasets using standard hardware together with the automatic optimization parameter selection.

We demonstrate this on several smaller datasets 
($10^4 \sim 10^5$ points) using a  Titan Xp GPU (see Table~\ref{tbl:small}).
We see that in every case training takes no more than $15$ seconds, making multiple runs for parameter and feature selection easily feasible.

For comparison, we also provide timings for LibSVM, a  popular and widely used kernel library~\cite{chang2011libsvm} and ThunderSVM~\cite{wen2018thundersvm}, a fast GPU implementation for LibSVM. We show the results for LibSVM~\footnote{We use the svm package in scikit-learn 0.19.0.} and ThunderSVM
using the same kernel with the same parameter. We stopped  iteration of our method when the accuracy on test exceeded that of LibSVM, which our method was able to achieve on every dataset. While not intended as a comprehensive evaluation, the benefits of our method for typical data analysis tasks are evident.\footnote{Our algorithm is still much faster than LibSVM when running on  CPU. For example, training on datasets shown in Table~\ref{tbl:small} takes between one and three  minutes. }
Fast training along with the ``worry-free'' optimization 
create an ``interactive/responsive'' environment for using kernel methods in machine learning.
Furthermore, the choice of kernel (e.g., Laplacian or Gaussian)  and its single bandwidth parameter is usually far simpler than the multiple parameters involved in the selection  of  architecture in neural networks.

\subsection{Practical Techniques for Accelerating Inference}

We would like to point out two simple and practical techniques to accelerate and simplify kernel training. The use of the Laplacian kernel is not common in the literature and in our opinion deserves more attention.  While PCA is frequently used to speed up training (and sometimes to improve the test results),  it is useful to state the technique explicitly.

\noindent {\bf Choice of kernel function.} In many cases Laplace (exponential)  kernel $k(\bx,\bz)=e^{-\frac{\|\bx-\bz\|}{\sigma}}$ produces results comparable or better than those for the more standard Gaussian kernel. Moreover the Laplacian kernel has several practical  advantages over the Gaussian (consistent with the findings reported in~\cite{belkin2018understand}).
(1) Laplacian generally requires fewer epochs for training to obtain the same quality result. 
(2) The batch value $m^*$ is typically larger for the Laplacian kernel allowing for more effective parallelization. (3) Test performance for the Laplacian kernel is empirically  more robust to the bandwidth parameter $\sigma$, significantly reducing the need for careful parameter tuning to achieve optimal performance.

\noindent {\bf Dimensionality reduction by PCA.}
Recall that the primary cost of one EigenPro iteration is $n \cdot md$ for the number of operations and $n\cdot (m + d)$ for memory where $d$ is the number of features. Thus reducing the dimension of the features results in significant computational savings. 
It is often possible to significantly reduce dimensionality of the data without perceptibly changing classification (or regression) accuracy by applying the Principal Components Analysis (PCA). 
For example, using PCA to reduce the feature dimensionality from $1536$ to $500$ for ImageNet decreases the accuracy by less than  $0.2\%$.

\section{Conclusion and Future Directions}

Best practices for training modern large-scale models 
are concerned with linear scaling. Most of the work is based on an implicit but widely held assumption that 
the limit of linear scaling, $m^*$, cannot be controlled or changed in practice.
In contrast, this paper shows that the limit of linear scaling can be analytically and automatically adapted to a given computing resource. This finding adds a new dimension for potential improvements in training large-scale models. 

The main technical contribution of this paper is a new learning framework (EigenPro 2.0) that 
extends linear scaling to match the parallel capacity of a computational resource. The framework is based on extracting limited second order information to modify the optimization procedure without changing the learned predictor function. While our paper deals with kernel machines, similar ideas are applicable to a much broader  class of learning architectures including deep neural networks. 

The algorithms developed in this paper allow for very fast kernel learning on smaller datasets and easy scaling to several million data points using a modern GPU.  It is likely that more effective memory management together with better hardware  would allow scaling up to $10^7$ data points with reasonable training time. Going beyond that to $10^8$ or  more data points using multi-GPU setups is the next natural step for kernel methods. 



\section*{Acknowledgements}
We thank Raef Bassily for discussions and helpful  comments and Alex Lee for running ThunderSVM comparisons. 
 We thank Lorenzo Rosasco and Luigi Carratino for sharing preprocessed ImageNet features. We used a Titan Xp GPU provided by Nvidia. We acknowledge financial support from NSF.


\bibliographystyle{sysml2019}
\bibliography{ref}

\newpage
\begin{appendices}
\section{Datasets}
We reduce multiclass labels to multiple binary labels.
For image datasets including MNIST~\cite{lecun1998gradient}, CIFAR-10~\cite{krizhevsky2009learning}, and SVHN~\cite{netzer2011reading}, color images are first transformed to grayscale images. We then rescale the range of each feature to $[0, 1]$.
For ImageNet~\cite{deng2009imagenet}, we use the top 800 PCA components of some convolutional features extracted from Inception-ResNet-v2~\cite{szegedy2017inception}.
For TIMIT~\cite{garofolo1993darpa}, we normalize each feature by z-score. 

\section{Selection of Kernel and its Bandwidth}
\begin{table*}[!ht]
\centering
\caption{Selected kernel bandwidth and corresponding optimization parameters}
\label{tbl:param}
\begin{adjustbox}{center}
\resizebox{\textwidth}{!}{
\begin{tabular}{|c|c||c|c||c|c|c|c|}
\hline
\multirow{2}{*}{Dataset} & \multirow{2}{*}{\begin{tabular}[c]{@{}c@{}}Size of (Subsampled)\\ Train Set\end{tabular}} & \multirow{2}{*}{Kernel} & \multirow{2}{*}{Bandwidth} & \multirow{2}{*}{Train epochs} & \multicolumn{3}{c|}{Calculated Parameters} \\ \cline{6-8} 
 &  &  &  &  & $q$ (adjusted $q$) & $m = m_\TM$ & $\eta$ \\ \hline
MNIST & $1 \cdot 10^6$ & Gaussian & 5 & 4 & 93 (330) & 735 & 379 \\ \hline
TIMIT & $1.1 \cdot 10^6$ & Laplacian & 15 & 3 & 52 (128) & 682 & 343 \\ \hline
ImageNet & $1.3 \cdot 10^6$ & Gaussian & 16 & 1 & 2 (321) & 294 & 149 \\ \hline
SUSY & $6 \cdot 10^5$ & Gaussian & 4 & 1 & 106 (850) & 1687 & 849 \\ \hline
\end{tabular}
}
\end{adjustbox}
\end{table*}
We use Gaussian kernel $k(x, y) = \exp(-\frac{\norm{x - y}^2}{2\sigma^2})$ and Laplace kernel $k(x, y) = \exp(-\frac{\norm{x - y}}{\sigma})$ in our experiments. Note that the kernel bandwidth $\sigma$ is selected through cross-validation on a small subsampled dataset.
In Table~\ref{tbl:param}, we report the kernel and its bandwidth selected for each dataset to achieve the best performance.
We also report the parameters that are calculated automatically using our method. Note that in practice we choose a value $q$ (in the parenthesis) that is larger than the $q$ corresponding to $m_\TM$. Increasing $q$ appears to lead to faster convergence. We use a simple heuristic to automatically obtain such $q$ based on the eigenvalue and the size of the fixed coordinate block\footnote{For SUSY  we directly specify a large $q$ for optimal performance.}.

\section{Analysis of Acceleration}
\label{sec:app-analysis}

\noindent {\bf Claim} (Acceleration). 
Using the adaptive kernel $\rmk_\TM$ decreases the resource time required for training over the original kernel $k$  by a factor of $a \approx \frac{\beta(K)}{\beta(K_\TM)} \cdot \frac{\mG}{m^*(k)}$.\\

We will now give a derivation of this acceleration factor $a$, based on the analysis of SGD in the interpolating setting in~\cite{ma2017interpolation}.  

As before, let  $(\bx_1, y_1), \ldots, (\bx_n, y_n)$ be the data, and let $K$ be the corresponding (normalized) kernel matrix $K_{ij}=k(\bx_i,\bx_j)/n$.
 We start by recalling the SGD iteration in the kernel setting for a mini-batch of size $m$,  $(\tx{1}, \ty{1}), \ldots, (\tx{m}, \ty{m})$,
$$
f \leftarrow f - \eta \cdot \frac{2}{m}
\left\lbrace \sum_{i=1}^m (f(\tx{i}) - \ty{i}) \rmk(\tx{i}, \cdot) \right\rbrace
$$

When step size $\eta$ is chosen optimally,
we can apply Theorem 4 in ~\cite{ma2017interpolation} to bound its convergence per iteration toward the optimal (interpolating) solution $f^*$ as follows:
$$
\expect{\norm{f_t - f^*}_\rmH^2}
\leq g^*_K(m) \cdot
\expect{\norm{f_{t-1} - f^*}_\rmH^2}
$$
Here $g_K^*(m)$ is a kernel-dependent upper bound on the convergence rate.

The fastest (up to a small constant factor) convergence rate  per iteration is obtained when using mini-batch size $m^*(K) =  \frac{\beta(K)}{\lambda_1(K)-\lambda_n(K)}$ (or larger). Kernels used in practice, such as Gaussian kernels, have rapid eigendecay~\cite{ma2017diving}, i.e., 
$\lambda_1(K)\gg\lambda_n(K)$. Hence we have 
$m^*(k) \approx  \frac{\beta(K)}{\lambda_1(K)}$.

Thus we can write an accurate approximation of convergence rate $g^*_K(m^*(K))$ as follows:
\begin{align*}
\epsilon^*_K \defeq g^*_K(m^*(K))
&= 1 - \frac{m^*(K) \cdot \lambda_n(K)}{\beta(K) + (m - 1) \lambda_n(K)}\\
& \approx 1 - \frac{\frac{\lambda_n(K)}{\lambda_1(K)}}{1+(m-1)\frac{\lambda_n(K)}{\beta(K)}}
\end{align*}
We now observe that $\beta= \max_{i=1,\ldots,n} k(\bx_i,\bx_i) \ge \tr(K)$.
Hence for  the mini-batch size $m$ much smaller than $n$ we have
$$
(m-1)\frac{\lambda_n(K)}{\beta(K)} \le (m-1)\frac{\lambda_n(K)}{\tr(K)} \le\frac{m-1}{n} \ll 1
$$

That allows us to write 
$$
\epsilon^*_K \approx 1 - \frac{\lambda_n(K)}{\lambda_1(K)}
$$

We will now apply this formula to the  adaptive kernel $k_\TM$. 
Recall that its corresponding kernel matrix $K_\TM$ modifies the top-$q$ eigenspectrum of $K$ such that
$$
\lambda_i(K_\TM) = \begin{cases}
\lambda_q(K) & \text{ if } i \leq q \\ 
\lambda_i(K) & \text{ if } i > q 
\end{cases} 
$$
Thus the convergence rate for $k_\TM$ is
$$
\epsilon^*_{K_\TM} \approx 1 - \frac{\lambda_n(K_\TM)}{\lambda_1(K_\TM)}
= 1 - \frac{\lambda_n(K)}{\lambda_q(K)}
$$



Next, we compare the number of iterations needed to converge to error $\epsilon$ using the original kernel $k$ and the adaptive kernel $k_\TM$.

First, we see that for kernel $k$ it takes $t= \frac{\log\epsilon}{\log\epsilon^*_K}$ iterations to go below error $\epsilon$ such that
$$\expect{\norm{f_t - f^*}_\rmH^2}
\leq \epsilon \cdot
\expect{\norm{f_{0} - f^*}_\rmH^2}$$

Notice that $\lambda_n(K) \leq \frac{\tr(K)}{n} = \frac{1}{n}$ for normalized kernel matrix $K$. Thus for large $n$, we have
$$
\frac{\log\epsilon}{\log\epsilon^*_K}
= \frac{\log\epsilon}{\log{(1 - \frac{\lambda_n(K)}{\lambda_1(K)})}}
\approx
\log\epsilon \cdot \frac{\lambda_1(K)}{\lambda_n(K)}
$$
In other words, the number of iterations needed to converge with kernel $k$ is proportional to
$\frac{\lambda_1(K)}{\lambda_n(K)}$.

By the same token,  to achieve accuracy $\epsilon$, the adaptive kernel $k_\TM$ needs
$
\frac{\log\epsilon}{\log\epsilon^*_{K_\TM}}
\approx
\log\epsilon \cdot \frac{\lambda_q(K)}{\lambda_n(K)}
$ iteration.

Therefore, to achieve accuracy $\epsilon$,
training with the adaptive kernel $k_\TM$ needs  
$ \frac{\lambda_q(K)}{\lambda_1(K)}$ as many iterations
 as training with the original kernel $k$.

To unpack the meaning of the ratio $\frac{\lambda_q(K)}{\lambda_1(K)}$, we rewrite it as
$$
\frac{\lambda_q(K)}{\lambda_1(K)} =
\frac{\lambda_1(K_\TM)}{\lambda_1(K)} 
= \frac{\beta(K_\TM)}{\beta(K)}
\cdot \frac{m^*(K)}{m^*(K_\TM)}=
\frac{\beta(K_\TM)}{\beta(K)} \cdot \frac{m^*(K)}{\mG}
$$

Recall that by the assumptions made in the paper (1)
any iteration for kernel $K$ with mini-batch size $m \leq \mG$ requires the same amount of resource time to complete on $\TM$,
(2) iteration of kernels $K$ and $K_\TM$ require the same resource time for any $m$ (negligible overhead). 

Since $m^*(K) \leq m^*(K_\TM) \approx \mG$,
we see that one iteration of batch size $m^*(K)$ and one iteration of batch size $m^*(K_\TM)$ take the same amount of time for either kernel. 

We thus conclude that the adaptive kernel accelerates over the original kernel by a factor of approximately
$$
\frac{\beta(K)}{\beta(K_\TM)} \cdot \frac{\mG}{m^*(K)}
$$

\paragraph{Remark.} Notice that our analysis is based on using upper bounds for convergence. While these bounds are tight (\cite{ma2017interpolation}, Theorem 3), there is no guarantees of tightness for  specific data and choice of kernel used in practice. Remarkably, the values of parameters obtained by using these bounds work very well in practice. Moreover, acceleration predicted theoretically  closely matches  acceleration observed in practice.


\end{appendices}

\end{document}